\def\isarxiv{1} 

\ifdefined\isarxiv
\documentclass[11pt]{article}
\else
\documentclass{article}
\usepackage{hyperref}
\usepackage[accepted]{icml2021}
\fi

\usepackage{microtype}
\usepackage{amsmath}
\usepackage{amsthm}
\allowdisplaybreaks
\usepackage{amssymb}
\usepackage{algorithm}
\usepackage{subfig}
\usepackage{color}
\usepackage{epstopdf}
\usepackage{url}
\usepackage{graphicx}
\usepackage{color}
\usepackage{epstopdf}
\usepackage{algpseudocode}
\usepackage{scrextend}
\usepackage[T1]{fontenc}
\usepackage{bbm}
\usepackage{comment}

\usepackage{multicol}
\usepackage{multirow}
\usepackage{dsfont}
\usepackage{mathtools}
\usepackage{enumitem}

\usepackage[makeroom]{cancel}
\usepackage{stmaryrd}
\usepackage{booktabs, makecell}
\usepackage{pifont}
\usepackage{lipsum}

\usepackage{tikz}

\usetikzlibrary{arrows}
\ifdefined\isarxiv
\usepackage{hyperref}
\hypersetup{colorlinks=true,citecolor=red,linkcolor=red}
\usepackage[margin=1in]{geometry}
\else

\fi
\graphicspath{{./figs/}}

\usepackage{amssymb}
\usepackage{latexsym}

\usepackage{url}
\usepackage{xcolor}
\usepackage{booktabs,caption}
\usepackage{threeparttable}
\usepackage{amsmath}
\usepackage{float}
\usepackage{graphicx}
\usepackage{placeins}
\usepackage{subfig}
\usepackage{tabularx}
\usepackage{natbib}
\usepackage{hyperref}
\usepackage{soul}
\usepackage{dashrule}
\usepackage{tikz}
\usepackage{siunitx}
\usepackage{bm}
\usepackage{diagbox}
\usepackage{fontawesome5}
\usepackage[table]{xcolor}

\newcommand{\ours}{$\mathtt{SiP}$}

\newcolumntype{M}{>{$}c<{$}}
\newcommand{\substd}[2]{#1_{\pm#2}}
\newcommand{\best}[1]{\ensuremath{\bm{#1}}}
\newcommand{\secondbest}[1]{\ensuremath{\underline{#1}}}

\definecolor{newcolor}{rgb}{.8,.349,.1}

\usepackage{amsmath,amsthm,amsfonts,bm}

\DeclareMathAlphabet{\mathsfit}{\encodingdefault}{\sfdefault}{m}{sl}
\SetMathAlphabet{\mathsfit}{bold}{\encodingdefault}{\sfdefault}{bx}{n}

\newcommand{\cX}{\mbox{$\mathcal{X}$}}

\renewcommand{\hat}{\widehat}

\newcommand{\vertiii}[1]{{\left\vert\kern-0.25ex\left\vert\kern-0.25ex\left\vert #1 
    \right\vert\kern-0.25ex\right\vert\kern-0.25ex\right\vert}}

\begin{document}

\title{See-in-Pairs: Reference Image-Guided Comparative Vision-Language Models for \\ Medical Diagnosis}
\date{}

\author{%
Ruinan Jin$^{1,2}$ \and
Gexin Huang$^{1,2}$ \and
Xinwei Shen$^{3}$ \and
Qiong Zhang$^{4}$ \and
Yan Shuo Tan$^{5}$ \and
Xiaoxiao Li$^{1,2}$\thanks{Corresponding author: Xiaoxiao Li (\texttt{xiaoxiao.li@ece.ubc.ca}).}%
}
\date{%
$^{1}$Electrical and Computer Engineering Department, The University of British Columbia, BC, Canada\\
$^{2}$Vector Institute, ON, Canada\\
$^{3}$ETH Zurich, R\"amistrasse 101, Zurich, Switzerland\\
$^{4}$Renmin University of China, Beijing, China\\
$^{5}$National University of Singapore, Queenstown, Singapore%
}



\begin{titlepage}
  \maketitle
  \begin{abstract}
  Medical image diagnosis is challenging because many diseases resemble normal anatomy and exhibit substantial interpatient variability. Clinicians routinely rely on comparative diagnostic, such as referencing \textit{cross-patient healthy control images} to identify subtle but clinically meaningful abnormalities. Although healthy reference images are abundant in practice, existing medical vision-language models (VLMs) primarily operate in a single-image or single-series setting and lack explicit mechanisms for comparative diagnosis. This work investigates whether incorporating clinically motivated comparison can enhance VLM performance. We show that providing VLMs with both a query image and a matched healthy reference image, accompanied by cross-patient comparative prompts, significantly improves diagnostic performance. This performance will be augmented by light-weight supervised finetuning (SFT) on small amount of data. At the same time, we evaluate \textit{multiple strategies for selecting reference images}, including random sampling, demographic attribute matching, embedding-based retrieval, and cross-center selection, and find consistently strong performance across all settings. Finally, we investigated how comparative diagnostic is effective theoretically, and observe improved sample efficiency and tighter alignment between visual and textual representations. Our findings highlight the clinical relevance of comparison-based diagnosis, provide practical strategies for incorporating reference images into VLMs, and demonstrate improved performance across diverse medical imaging tasks.
  \end{abstract}
  \thispagestyle{empty}
\end{titlepage}



\section{Introduction}
\label{sec:intro}
Medical image diagnosis is a central application of medical AI, spanning modalities such as Chest Xray, CT, MRI, ultrasound, and fundus imaging~\citep{moor2023foundation,sun2025visual,koccak2025ai}. Yet, unlike natural images, where categories are visually distinct, pathological findings often appear as subtle and highly localized deviations embedded within large amounts of normal anatomy, with substantial inter-patient variability~\citep{shen2017deep,anwar2018medical}. Such characteristics challenge the current medical AI, which are not explicitly optimized for fine-grained medical cues~\citep{razzak2017deep,chan2020deep}.

\begin{figure}
    \centering
    \includegraphics[width=0.5\linewidth]{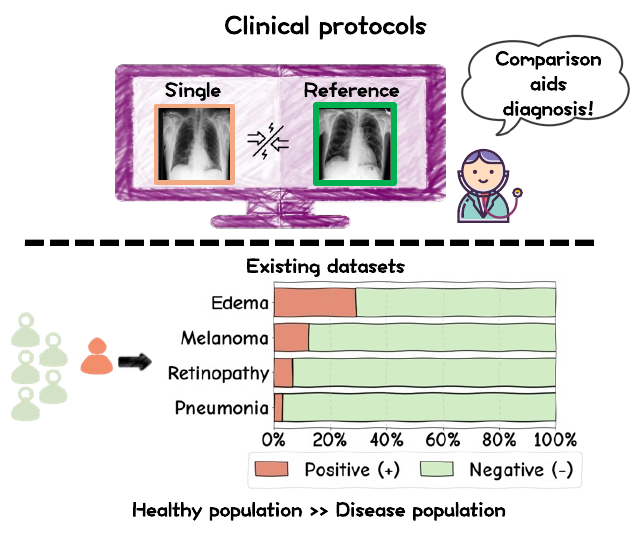}
    \caption{\textbf{Clinical motivation of this paper}: clinicians leverage healthy-control reference images to aid diagnosis, yet existing datasets are dominated by healthy images.}
    \label{fig:motivation}
\end{figure}

As shown in Fig.~\ref{fig:motivation}, in clinical practice, clinicians routinely perform \textit{comparative diagnosis}: juxtaposing the query image with similar examinations or healthy controls to isolate subtle abnormalities~\citep{lange2022influence,kok2015case,kok2013learning,sukut2023providing}. Controlled clinical studies confirm that pairing abnormal and normal images significantly improves diagnostic performance~\citep{kok2015case,sukut2023providing}. 

These observations motivate our central question: \ul{\textit{Can clinically inspired cross-subject comparison help medical AI better utilize abundant healthy control images and thereby improve diagnostic performance?}}


Recent progress in Vision Language Models (VLMs) has demonstrated a promising ability to understand multiple images within the natural image domain~\citep{meng2024mmiu,zhao2024benchmarking, wahed2024prima}. Models such as QwenVL-2.5~\citep{bai2025qwen2}, NVILA~\citep{liu2024nvila}, and Phi-3.5~\citep{abdin2024phi} can effectively identify and articulate fine-grained semantic differences across various images or viewpoints. However, their lack of medical knowledge limits their clinical reliability and interpretability. Medical VLMs~\citep{de2025padchest,chen2024huatuogpt,li2023llava,thawkar2023xraygpt,busch2025large} are developed to integrate medical image and textual data (e.g., clinical reports) for enhanced clinical decision-making, though the current medical VLMs like LLaVa-Med~\citep{li2023llava} and XrayGPT~\citep{thawkar2023xraygpt} primarily focus on analyzing single images and are not optimized for comparative diagnosis across different subjects. This limitation largely stems from the instructional tuning of medical datasets, such as PMC figure–caption pairs~\citep{li2023llava}, which typically follow a single-image diagnosis format. While recent work~\citep{yang2025medical} has begun to explore VLMs for multi-image tasks, it mainly concentrates on intra-patient image-caption data, spanning multiple visits, views, or modalities, to assess a single patient's progression (e.g., comparing previous and current X-rays) and co-reference: identifying which image contains specific content. Crucially, existing research focuses on analyzing images from a single patient rather than facilitating comparisons between different subjects. This oversight ignores a common diagnostic scenario where only a single static image is available per patient, leaving a significant gap in our ability to perform comparative analysis across different subjects to gain broader diagnostic insights.

In response to the scarcity of methods supporting comparative diagnosis across subjects, this study explores two potential solutions. \textit{First}, we assess off-the-shelf VLMs with their zero-shot feasibility, finding that general-purpose models with multi-image priors respond well to structured (query, reference) prompting (Sec.~\ref{exp:off-the-shelf}). However, the failure of this approach in standard medical VLMs highlights a critical gap: specialized models require explicit training to develop comparative capability, as mere reference injection is inadequate for models lacking \textit{cross-subject} pretraining. Given the prohibitive cost of training such cross-subject understanding capacity~\citep{meng2024mmiu,yang2025medical}, we introduce a lightweight supervised fine-tuning (SFT) strategy and practical fine-tuning data construction methods that formulate (query, reference, label) triplets with clinically meaningful negatives. Our method uses limited labeled data to adapt general-purpose VLMs with medical knowledge for cross-subject comparative analysis and significantly improves performance on medical diagnosis. Our subsequent analysis validates the utility of this paradigm; both theoretical bounds and representational visualizations confirm that comparative analysis helps VLMs make the model to focus on pathology-specific deviations while remaining robust to normal anatomical variation.

Our key contributions are:
\begin{itemize}
\item \textbf{New perspective}: We identify \textit{cross-subject comparative diagnosis} as an essential yet overlooked direction for medical VLMs. We argue that models should mimic the clinical practice of comparing patient scans against reference cases.
\item \textbf{Zero-Shot feasibility}: We reveal that general-purpose VLMs capable of multi-image comparison outperform single-image queries in zero-shot medical diagnosis when using structured (query, reference) inputs. This confirms that comparative prompting is a viable strategy, even without specific training.
\item \textbf{Scalable SFT framework}: We propose a lightweight SFT framework and a practical data construction method to improve general-purpose VLM models that are capable of cross-patient comparison. By using (query, reference, label) triplets and clinically inspired negative references, we effectively inject comparative medical knowledge into these general-purpose VLM models.
\item \textbf{Robust performance}: We demonstrate consistent improvements in diagnosis across six medical datasets. Our results hold up across various reference-selection strategies, including random sampling, demographic matching, and cross-center testing.
\item \textbf{Mechanistic insight}: We provide theoretical and empirical evidence explaining underlying mechanisms of how comparative diagnostic improves VLM performance, linking it to stronger feature alignment and reduced nuisance variation (like scanner differences).
\end{itemize}

\section{Related Work}
\begin{table*}[h]
\centering
\small
\setlength{\tabcolsep}{3pt}
\caption{\textbf{Summary of existing VLMs.} Comparison of recent VLMs in terms of their architectural and data characteristics. Columns indicate whether each model incorporates a \emph{cross-image comparison prior}, includes a \emph{medical prior}, and provides \emph{open-source weights}. Green check marks denote presence, while red crosses indicate absence of the respective property.}
\begin{tabular}{lccc}
\toprule
Model & Cross-subject Comp. Prior & Medical Prior & Open-Weights \\
\midrule
BLIP2~\citep{li2023blip} 
  & \textcolor{red}{\faTimes} 
  & \textcolor{red}{\faTimes} 
  & \textcolor{green}{\faCheck} \\
InstructBLIP~\citep{dai2023instructblip} 
  & \textcolor{red}{\faTimes} 
  & \textcolor{red}{\faTimes} 
  & \textcolor{green}{\faCheck} \\
LLaVa~\citep{liu2023visual} 
  & \textcolor{red}{\faTimes} 
  & \textcolor{red}{\faTimes} 
  & \textcolor{green}{\faCheck} \\
QwenVL-2.5-Instruct~\citep{bai2025qwen2} 
  & \textcolor{green}{\faCheck} 
  & \textcolor{red}{\faTimes} 
  & \textcolor{green}{\faCheck} \\
Phi-3.5-Vision~\citep{abdin2024phi} 
  & \textcolor{green}{\faCheck} 
  & \textcolor{red}{\faTimes} 
  & \textcolor{green}{\faCheck} \\
R2Gen~\citep{chen2020generating} 
  & \textcolor{red}{\faTimes} 
  & \textcolor{green}{\faCheck} 
  & \textcolor{green}{\faCheck} \\
LLaVaMed~\citep{li2023llava} 
  & \textcolor{red}{\faTimes} 
  & \textcolor{green}{\faCheck} 
  & \textcolor{green}{\faCheck} \\
XrayGPT~\citep{thawkar2023xraygpt} 
  & \textcolor{red}{\faTimes} 
  & \textcolor{green}{\faCheck} 
  & \textcolor{green}{\faCheck} \\
MedVLM-R1~\citep{pan2025medvlm} 
  & \textcolor{red}{\faTimes} 
  & \textcolor{green}{\faCheck} 
  & \textcolor{green}{\faCheck} \\
MUMC~\citep{li2023masked} 
  & \textcolor{red}{\faTimes} 
  & \textcolor{green}{\faCheck} 
  & \textcolor{green}{\faCheck} \\
MedFlamingo~\citep{moor2023med}\footnotemark 
  & \textcolor{green}{\faCheck} 
  & \textcolor{green}{\faCheck} 
  & \textcolor{red}{\faTimes} \\
NVILA~\citep{liu2024nvila} 
  & \textcolor{green}{\faCheck} 
  & \textcolor{green}{\faCheck} 
  & \textcolor{green}{\faCheck} \\
HuatuoGPT-Vision~\citep{chen2024huatuogpt} 
  & \textcolor{red}{\faTimes} 
  & \textcolor{green}{\faCheck} 
  & \textcolor{green}{\faCheck} \\
\bottomrule
\end{tabular}
\label{tab:vlm}
\end{table*}

\footnotetext{The checkpoint was inaccessible during preparation of this paper as shown by this \href{https://github.com/snap-stanford/med-flamingo/issues/18}{issue}.}

\paragraph{Comparative diagnosis in clinics} Comparative diagnosis, the process of evaluating clinical findings through reference to other medical images, is a pervasive principle across medical practice. In radiology, comparing a current study with prior examinations or healthy control reference images is known to improve diagnostic accuracy and is embedded in clinical guidelines~\citep{lange2022influence,kok2013learning}. This comparative diagnostic extends beyond radiology and single-series images: in pathology, cross-slide comparison between normal and diseased tissue is fundamental for recognizing subtle morphological deviations~\citep{wicks2023comparative}; in dermatology, clinicians routinely compare lesions with known benign exemplars or contralateral body sites~\citep{cornell2015viewing}; and in ophthalmology or endoscopy, side-by-side visualization of normal reference images enhances anomaly detection~\citep{cavichini2023accuracy}. Furthermore, clinical studies further show that juxtaposing normal and abnormal exemplars improves diagnostic discrimination and perceptual learning across visual specialties~\citep{sukut2023providing,kok2013learning,kok2015case,eder2021comparing}, making cross-subject (e.g., abnormal vs. normal reference) a universal diagnostic strategy.


\paragraph{VLM with multi-image comparison capability}
Recent general-purpose VLMs have increasingly incorporated capabilities for image comparison by leveraging architectural advances and multimodal training strategies. In this work, we focus on open-source VLMs to ensure our findings are reproducible and mechanistically understandable. For example, Qwen-VL-2.5-Instruct~\citep{bai2025qwen2} facilitates fine-grained visual comparison by utilizing co-attention-based multimodal fusion, image-specific positional encoding, and instruction tuning designed for comparative reasoning tasks. NVILA-8B~\citep{liu2024nvila} employs a ``scale-then-compress'' architecture that allows high-resolution and long-context visual input, supporting nuanced relational reasoning across images. 
Phi-3.5-Vision~\citep{abdin2024phi} introduces interleaved visual-text token processing, facilitating multi-hop relational inference across multiple images in a single context window. However, not all general-purpose VLMs possess this comparative capability; for example, BLIP2~\citep{li2023blip} and LLaVA~\citep{li2023llava} lack this feature. In the medical domain, the bulk of recent medical VLMs have been developed for single‐image or single‐series (e.g., longitudinal studies from one patient) understanding~\citep{chen2020generating,li2023llava,thawkar2023xraygpt,pan2025medvlm,li2023masked,cho2024pretraining}, with comparatively little emphasis on cross‐subject image comparison. Table~\ref{tab:vlm} lists the major open-source general-purpose and medical VLMs on three perspectives \

\begin{figure*}[t]
    \centering
    \includegraphics[width=0.95\linewidth]{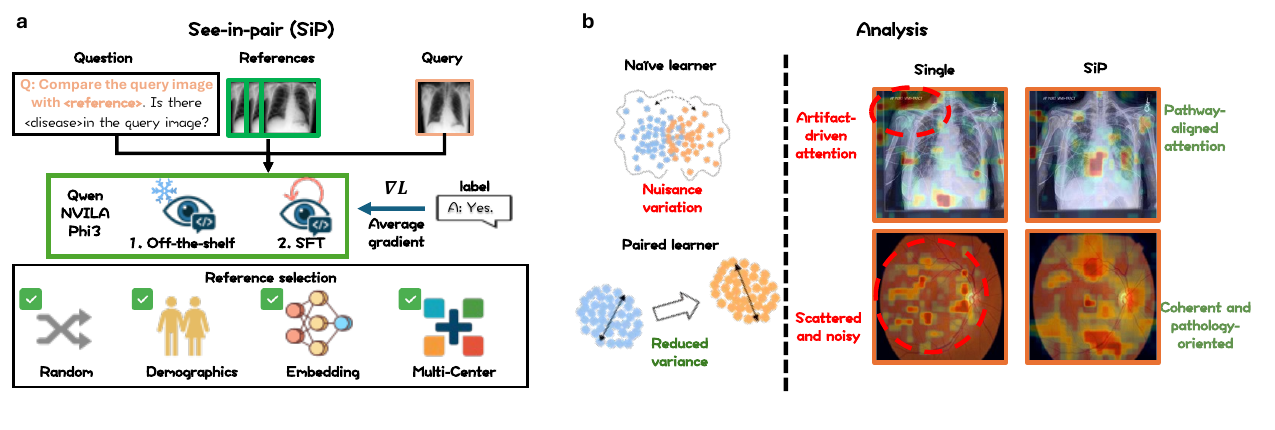}
    \caption{\textbf{Overview of our study pipeline.} 
(\textbf{a}) \emph{\ours{} method overview}: we first explored the off-the-shelf inference performance when leveraging healthy-control reference images and then conducted comparative SFT by constructing (query, reference, label) triples and fine-tuning the VLMs (e.g., Qwen, NVILA, Phi-3). We additionally study several clinically inspired reference-selection strategies, including random sampling, demographic matching, embedding-based retrieval, and cross-center sampling. 
(\textbf{b}) \emph{Analysis}: pairing query images with reference images reduces nuisance variation and improves feature alignment, enabling more reliable diagnostic performance compared to a single-image setting.}
\label{fig:main}
\end{figure*}

\section{See-in-Pairs}
\label{sec:method}
To address the research question introduced in Sec.~\ref{sec:intro}, we begin by evaluating the zero-shot performance of the model, comparing settings \emph{with} and \emph{without} comparative diagnosis (Sec.~\ref{method:inference}). Next, we examine whether incorporating comparative diagnosis improves performance during supervised fine-tuning (Sec.~\ref{method:SFT}). Finally, we analyze the underlying mechanisms that explain why \ours{} yields performance gains. Notations used in Sec.~\ref{sec:method} and Sec.~\ref{sec:theory} are summarized in the Table~\ref{tab:notation}.



\subsection{\ours{} Inference}
\label{method:inference}
\paragraph{Single query image inference} Let $\cX$ be the space of medical images and $\mathcal{Q}$ the space of natural-language prompts.  
We draw triplets $(X,Q,A)$, where $X\in\cX$, $Q\in\mathcal{Q}$, and $A=\{a_1,\dots,a_K\}$ is the potential ground-truth answer list, where $a_k$ could be ``yes'', ``no'', or disease types. We serialize the $(X, Q)$ pair as the input prompts to the LLM:
\begin{equation}\label{eq:x}
Z = \bigl\langle 
         \operatorname{encoder}(X),\,
         \operatorname{encode}(Q)
    \bigr\rangle .
\end{equation}
Here $\operatorname{encode}(X)$ denotes patch embeddings from a vision encoder. For any candidate answer string,
$a_k=\langle t_{1},\dots,t_{L_k}\rangle$, where $t_i$ is the $i^{th}$ token and $L_k$ is the maximum length of the given answer. We make a prediction via a VLM parameterized by $\theta$,
\begin{equation}\label{eq:inference}
\hat a
=
\arg\max_{a_k\in\mathcal A}
P_\theta(a_k\mid Z),\qquad
P_\theta(a_k\mid Z)=
\prod_{j=1}^{L_k}P_\theta\!\bigl(t_j\mid Z,t_{<j}\bigr).
\end{equation} for the final prediction. 

\paragraph{Inference with reference image} To mimic clinical practice, we introduce a reference image $X' \in \cX'$ with known labels. In reference-based medical diagnosis, often in the format of anomaly detection, the reference standard is typically established using non-pathological baselines. This approach is critical for defining a robust normative distribution, ensuring that pathological deviations are detected with high precision. Additionally, this strategy addresses the data imbalance inherent in medical imaging, leveraging common healthy control data compared to the scarcity of comprehensive, disease-specific datasets, thereby ensuring a consistent, universal, and scalable reference protocol regardless of the specific pathology. Therefore, in our formulation, the reference images are assigned negative diagnostic labels, such as ``no finding'' or ``healthy control.'' The primary purpose of incorporating reference images is to help the VLM focus on disease–related visual patterns while reducing the influence of irrelevant factors, such as scanner variability or patient attributes like gender and age.

We pair each query image with one or more negative reference images using five strategies (results shown in Sec.~\ref{exp:selection}):
\begin{itemize}
    \item \textit{Random sampling}: for each query image, we uniformly sample a single reference image from the pool of healthy-control images in the training set, excluding the query image itself. This strategy provides a simple and computationally lightweight baseline that does not rely on metadata or similarity metrics.
    
    \item \textit{Demographic matching}: for each query image, we select a single healthy reference image whose available patient- or acquisition-level attributes (e.g., gender, imaging view, or projection) match those of the query. This strategy reflects clinically meaningful comparisons but is limited to datasets with sufficient demographic or acquisition annotations.
    
    \item \textit{Embedding-based retrieval}: for each query image, we select a single healthy reference image that is most similar in feature space, using precomputed image embeddings from a pretrained encoder. This approach enforces semantic relevance between the query and reference without requiring explicit metadata.
    
    \item \textit{Cross-center sampling}: for each query image, we sample reference images exclusively from a different medical center or dataset (e.g., using MIMIC-CXR~\citep{johnson2019mimic} images as references for CheXpert~\citep{irvin2019chexpert} queries). This strategy evaluates the robustness of \ours{} to domain shifts in acquisition protocols and patient populations.
    
    \item \textit{Bagging}: instead of selecting a single reference image, we select multiple reference images per query using random sampling. The model is evaluated independently on each (query, reference) pair, and the final prediction is obtained by majority voting over the resulting outputs. This ensemble-style aggregation reduces sensitivity to individual reference choices. 
\end{itemize}

Unlike single-image inference, we serialize the $(X, X', Q)$ triplet as the input prompts to the VLM: 
\begin{equation}\label{eq:x'}
Z' = \bigl\langle \operatorname{fusion}(
         \operatorname{encode}(X),\, 
         \operatorname{encode}(X')),\,
         \operatorname{encode}(Q)
    \bigr\rangle,
\end{equation}
where $\operatorname{fusion}$ is a VLM-dependent fusion operation and then we replace $Z$ with $Z'$ in Eq.~\ref{eq:inference}.

\subsection{Improved Performance via Comparative Supervised Fine-tuning}
\label{method:SFT}
While the results of zero-shot testing presented in Table.~\ref{tab:combined-zs} demonstrate the promise of incorporating reference images into medical imaging diagnosis, we also observe a clear performance gap when relying solely on zero-shot inference using general-purpose VLMs. Following established practices for SFT in generative medical VLMs~\citep{li2023llava}, we focus our updates specifically on the language decoder component. Accordingly, we decompose the \texttt{VLM} = \texttt{LE} $\circ$ \texttt{VE} into a fixed vision expert (\texttt{VE}) and a learnable language expert (\texttt{LE}) (\textit{i.e.,} LLM decoder) denoted as function $g_\theta$. For simplicity and clarity in this method, the multi-image fusion and projection layers are implicitly included within \texttt{VE}. We first define SFT on single image then show the SFT on inputs with reference images.

\noindent\textbf{Comparative SFT with reference images (\ours{}).}
For completeness, we first recall the standard single-image SFT objective used by our single-image baselines. For ground-truth answer string
$a^*=\langle t_{1},\dots,t_{L}\rangle$, the generative VLM is optimized on $(X,Q,A)$ via
\begin{align}
\mathcal{L}_{\rm SFT}(\theta) 
&= -\sum_{x}\sum_{j} \log p_{\theta}(t_j|x,t_{<j}),
    && \text{where} \nonumber \\
p_{\theta}(t_j|x,t_{<j}) 
&= \frac{\exp\!\big(g_{\theta}(x,t_{<j})_{t_j}\big)}
        {\sum_{t'\in\mathcal{V}} 
         \exp\!\big(g_{\theta}(x,t_{<j})_{t'}\big)} .
\label{eq:single_sft}
\end{align}
where $x$ is the input to \texttt{LE} as defined in Eq.~\ref{eq:x}.
Our comparative SFT constructs inputs as $(X, X', Q, A)$, wherein $X$-matched reference images $X'$ are sampled using the strategy described before, specifically negative images that share diagnosis-irrelevant features. The objective is identical to Eq.~\ref{eq:single_sft} but with $x$ replaced by $Z'$, where $Z'$ is defined in Eq.~\ref{eq:x'}. Furthermore, during training, multiple (say $K$) $X'$ instances can be sampled for a given $X$ to form various SFT tuples, and the model can be updated using the averaged gradient across these tuples.\footnote{This is implemented by scaling the number of gradient accumulation steps in proportion to the number of reference images sampled per query. An ablation study illustrating this effect is provided in the Appendix.} Similarly, at inference time, we can sample multiple reference images for the same query and re-prompt the VLM for each sampled pair; we refer to this as \emph{bagging}. The resulting predictions are then aggregated (e.g., by majority vote) to further improve robustness and performance.
\section{Experiments}
\label{sec:exp}
\subsection{Settings}
\label{exp:setting}

\paragraph{Datasets}
To ensure a comprehensive evaluation, we assess \ours{} across a broad spectrum of medical imaging data. Our study spans four distinct imaging modalities: \textit{radiology}, \textit{OCT retinal nerve fiber layer (RNFL) thickness maps}, \textit{dermatoscopic images}, and \textit{color fundus photography}, capturing diversity in visual characteristics and clinical use cases. The datasets additionally vary substantially in scale, ranging from large radiology cohorts to small OCT collections, and cover both binary and multi-class diagnostic tasks. 

For radiology, we evaluate two diagnostic tasks, \texttt{Pneumonia} and \texttt{Edema}\footnote{ Each \texttt{<disease>} token denotes a specific diagnostic task and is used consistently throughout all tables and figures.}, collected from the CheXpert dataset~\citep{irvin2019chexpert}. In both tasks, images labelled as ``No Finding'' are treated as \emph{healthy controls} and form the reference pool. For OCT RNFL imaging of the retina, we perform \texttt{Glaucoma} diagnosis collected from the GF3300 dataset~\citep{luo2024harvard}. Eyes graded as ``normal'' constitute the healthy-control set used as references. For dermatology, we employ the HAM10000 dataset~\citep{tschandl2018ham10000}, conducting both a binary \texttt{Melanoma} diagnostic task and a multi-class task, \texttt{DermaTri}, which covers \textit{Basal Cell Carcinoma}, \textit{Melanoma}, and \textit{Melanocytic}. Benign skin lesions were used as healthy controls for our study. Finally, for colorful ophthalmology, we collected data from the BRSET dataset~\citep{nakayama2024brset} to evaluate \texttt{Retinopathy} diagnosis, where grade~0 images (no retinopathy) serve as healthy controls. Table~\ref{tab:test_split} shows the detailed statistics of each task.

\begin{table}[h!]
    \centering
    \captionsetup{width=\columnwidth}
    \caption{\textbf{Class distribution in the testing splits.} For \texttt{DermaTri}, the two positive classes contain 103 samples each (total \num{206}).}
    \label{tab:test_split}
    \small
    \renewcommand{\arraystretch}{1.15}
    \begin{tabular*}{\columnwidth}{@{\extracolsep{\fill}} l S[table-format=4.0] S[table-format=4.0] @{}}
        \toprule
        \textbf{Task} & {\textbf{Test +}} & {\textbf{Test --}} \\
        \midrule
        \texttt{Edema}       & 4000 & 4000 \\
        \texttt{Pneumonia}   & 2000 & 2000 \\
        \texttt{Glaucoma}    & 303  & 291  \\
        \texttt{Melanoma}    & 500  & 500  \\
        \texttt{DermaTri}    & 206  & 103  \\
        \texttt{Retinopathy} & 556  & 556  \\
        \bottomrule
    \end{tabular*}
\end{table}

\paragraph{Models} 
As mentioned in the Sec.~\ref{sec:intro}, we focus on VLMs that both exhibit strong comparative capabilities and provide open-source fine-tuning code to facilitate lightweight adaptation. For assessing comparison capabilities, we refer to the benchmark proposed by~\cite{zhao2024benchmarking}, which offers a comprehensive evaluation of existing VLMs on general-domain comparative tasks. We require selected models to support multi-image inference and, ideally, to incorporate medical domain priors for improved off-the-shelf performance. Among the benchmark results~\citep{liu2024nvila}, Phi-3-4B and NVILA-7B demonstrate strong multi-image reasoning capabilities, which we decided to include into our study. Also, we include the QwenVL-7B~\citep{bai2025qwen2} as it exhibits good generalization performance and is also adapted in various medical VLM studies this year~\citep{lai2025med,zhong2025can}.

\begin{table*}[h!]
\centering
\footnotesize  
\setlength{\tabcolsep}{3.5pt} 
\caption{\textbf{Off-the-shelf performance of general-purpose and medical VLMs on diagnostic tasks.} Performance (BAcc. and F1, \%) is shown for two settings: Single-image (\textbf{S}), and Comparative \ours (\textbf{C}). The tasks are diagnosing \texttt{Pneumonia}, \texttt{Edema}, \texttt{Glaucoma}, \texttt{Melanoma}, \texttt{DermaTri}, and \texttt{Retinopathy}. Values are reported as mean~$\pm$~standard deviation. The best result between S and C settings is highlighted in \textbf{bold} for each model and dataset. We note that many diagnostic tasks considered here are out-of-distribution for current VLMs, which are largely trained without reinforcement learning or large-scale multimodal instruction tuning, leading to limited generalization and potentially lower off-the-shelf performance compared to general-purpose VLMs. This observation is aligned with~\cite{zhong2025can,xu2025lingshu}.}
\label{tab:combined-zs}
\resizebox{0.99\linewidth}{!}{%
\begin{tabular}{l c | cc | cc | cc | cc | cc | cc}
    \toprule
    \multirow{2}{*}{\textbf{Model}} & \multirow{2}{*}{} 
      & \multicolumn{2}{c|}{\textbf{Pneumonia}} 
      & \multicolumn{2}{c|}{\textbf{Edema}} 
      & \multicolumn{2}{c|}{\textbf{Glaucoma}} 
      & \multicolumn{2}{c|}{\textbf{Melanoma}} 
      & \multicolumn{2}{c|}{\textbf{DermaTri}} 
      & \multicolumn{2}{c}{\textbf{Retinopathy}} \\
    \cmidrule(lr){3-4} \cmidrule(lr){5-6} \cmidrule(lr){7-8} \cmidrule(lr){9-10} \cmidrule(lr){11-12} \cmidrule(lr){13-14}
      & & \textbf{BAcc} & \textbf{F1} & \textbf{BAcc} & \textbf{F1} & \textbf{BAcc} & \textbf{F1} & \textbf{BAcc} & \textbf{F1} & \textbf{BAcc} & \textbf{F1} & \textbf{BAcc} & \textbf{F1} \\
    \midrule
    \multicolumn{14}{l}{\textit{General-purpose VLMs}} \\
    \midrule
    \multirow{2}{*}{QwenVL} 
      & S & $\substd{55.30}{0.77}$ & \best{\substd{50.23}{1.04}} & $\substd{50.14}{0.05}$ & $\substd{0.64}{0.17}$ & \best{\substd{53.03}{1.94}} & $\substd{44.98}{2.65}$ & \best{\substd{55.55}{1.29}} & \best{\substd{65.36}{1.53}} & $\substd{41.30}{2.25}$ & $\substd{32.77}{2.03}$ & $\substd{52.43}{0.65}$ & $\substd{13.95}{1.88}$ \\
      & C & \best{\substd{58.35}{0.65}} & $\substd{43.70}{1.10}$ & \best{\substd{52.79}{0.43}} & \best{\substd{29.64}{0.81}} & $\substd{52.95}{1.99}$ & \best{\substd{54.28}{2.39}} & $\substd{51.01}{1.25}$ & $\substd{62.73}{1.60}$ & \best{\substd{42.93}{2.22}} & \best{\substd{34.22}{2.00}} & \best{\substd{55.08}{1.41}} & \best{\substd{46.09}{1.97}} \\
    \cmidrule{1-14}
    \multirow{2}{*}{\quad + bagging} 
      & S & $\substd{56.23}{0.77}$ & \best{\substd{51.16}{1.00}} & $\substd{50.14}{0.05}$ & $\substd{0.65}{0.18}$ & $\substd{52.32}{1.94}$ & $\substd{44.23}{2.72}$ & \best{\substd{56.10}{1.23}} & \best{\substd{66.05}{1.52}} & $\substd{42.39}{2.21}$ & $\substd{34.16}{2.11}$ & $\substd{52.52}{0.65}$ & $\substd{14.01}{1.90}$ \\
      & C & \best{\substd{59.25}{0.63}} & $\substd{42.97}{1.18}$ & \best{\substd{53.16}{0.55}} & \best{\substd{24.47}{0.79}} & \best{\substd{55.85}{2.10}} & \best{\substd{57.19}{2.38}} & $\substd{49.50}{1.10}$ & $\substd{62.84}{1.51}$ & \best{\substd{44.98}{2.09}} & \best{\substd{35.82}{1.90}} & \best{\substd{56.74}{1.32}} & \best{\substd{44.78}{2.06}} \\
    \midrule
    \multirow{2}{*}{Phi} 
      & S & $\substd{51.68}{0.47}$ & $\substd{19.93}{1.05}$ & $\substd{49.90}{0.19}$ & $\substd{5.60}{0.49}$ & $\substd{51.48}{0.71}$ & $\substd{8.74}{2.10}$ & $\substd{49.98}{0.60}$ & $\substd{6.74}{1.53}$ & \best{\substd{36.53}{1.68}} & \best{\substd{27.37}{2.58}} & $\substd{50.73}{0.65}$ & $\substd{11.02}{1.72}$ \\
      & C & \best{\substd{54.68}{0.68}} & \best{\substd{63.52}{0.82}} & \best{\substd{50.69}{0.42}} & \best{\substd{62.87}{0.53}} & \best{\substd{51.75}{1.83}} & \best{\substd{61.65}{2.15}} & \best{\substd{51.68}{1.22}} & \best{\substd{63.26}{1.52}} & $\substd{33.97}{0.45}$ & $\substd{18.11}{1.30}$ & \best{\substd{58.95}{1.43}} & \best{\substd{63.25}{1.63}} \\
    \cmidrule{1-14}
    \multirow{2}{*}{\quad + bagging} 
      & S & $\substd{51.70}{0.50}$ & $\substd{19.97}{1.10}$ & $\substd{49.89}{0.20}$ & $\substd{5.60}{0.49}$ & $
      \substd{51.29}{0.70}$ & $\substd{8.72}{2.12}$ & $\substd{50.00}{0.60}$ & $\substd{6.72}{1.51}$ & \best{\substd{36.57}{1.67}} & \best{\substd{27.53}{2.55}} & $\substd{50.90}{0.65}$ & $\substd{11.36}{1.72}$ \\
      & C & \best{\substd{55.55}{0.64}} & \best{\substd{65.45}{0.77}} & \best{\substd{50.25}{0.37}} & \best{\substd{63.90}{0.53}} & \best{\substd{52.23}{1.54}} & \best{\substd{64.38}{1.91}} & \best{\substd{50.99}{1.10}} & \best{\substd{63.92}{1.56}} & $\substd{33.98}{0.47}$ & $\substd{18.14}{1.39}$ & \best{\substd{58.99}{1.41}} & \best{\substd{63.17}{1.61}} \\
    \midrule
    \multicolumn{14}{l}{\textit{Medical VLMs}} \\
    \midrule
    \multirow{2}{*}{NVILA} 
      & S & \best{\substd{69.31}{0.66}} & $\substd{61.79}{1.00}$ & \best{\substd{76.72}{0.43}} & $\substd{71.77}{0.62}$ & $\substd{50.15}{0.30}$ & $\substd{1.29}{0.91}$ & $\substd{44.87}{1.54}$ & $\substd{47.97}{1.90}$ & $\substd{37.41}{1.71}$ & $\substd{27.19}{2.04}$ & $\substd{53.42}{0.74}$ & $\substd{18.30}{2.06}$ \\
      & C & $\substd{61.80}{0.67}$ & \best{\substd{68.59}{0.77}} & $\substd{73.46}{0.49}$ & \best{\substd{72.20}{0.59}} & \best{\substd{53.82}{2.00}} & \best{\substd{56.32}{2.30}} & \best{\substd{52.11}{1.04}} & \best{\substd{64.97}{1.48}} & \best{\substd{39.11}{1.42}} & \best{\substd{27.80}{2.12}} & \best{\substd{56.30}{1.41}} & \best{\substd{61.97}{1.58}} \\
    \cmidrule{1-14}
    \multirow{2}{*}{\quad + bagging} 
      & S & \best{\substd{69.34}{0.66}} & $\substd{61.83}{1.00}$ & \best{\substd{76.65}{0.42}} & $\substd{71.70}{0.62}$ & $\substd{49.99}{0.25}$ & $\substd{0.66}{0.65}$ & $\substd{44.90}{1.53}$ & $\substd{48.26}{1.83}$ & $\substd{37.54}{1.69}$ & $\substd{27.40}{2.02}$ & $\substd{53.42}{0.76}$ & $\substd{18.30}{2.10}$ \\
      & C & $\substd{61.85}{0.80}$ & \best{\substd{69.36}{0.77}} & $\substd{76.60}{0.46}$ & \best{\substd{74.82}{0.55}} & \best{\substd{54.07}{2.01}} & \best{\substd{57.50}{2.29}} & \best{\substd{51.30}{1.52}} & \best{\substd{65.68}{1.40}} & \best{\substd{39.16}{1.29}} & \best{\substd{27.68}{2.06}} & \best{\substd{56.29}{1.48}} & \best{\substd{61.97}{1.59}} \\
    \midrule
    \multirow{2}{*}{HuatuoVision} 
      & S & \best{\substd{60.16}{0.67}} & $\substd{47.47}{1.14}$ & $\substd{56.85}{0.37}$ & $\substd{31.87}{0.84}$ & $\substd{50.50}{1.19}$ & $\substd{15.77}{2.63}$ & $\substd{54.00}{1.28}$ & $\substd{64.78}{1.55}$ & $\substd{35.26}{0.81}$ & $\substd{20.55}{1.76}$ & $\substd{49.71}{0.97}$ & $\substd{19.30}{2.01}$ \\
      & C & $\substd{54.38}{0.78}$ & \best{\substd{59.99}{0.89}} & \best{\substd{58.77}{0.56}} & \best{\substd{54.10}{0.73}} & \best{\substd{51.48}{1.92}} & \best{\substd{56.45}{2.26}} & $\substd{50.50}{0.97}$ & $\substd{64.62}{1.49}$ & \best{\substd{38.16}{1.15}} & \best{\substd{25.44}{2.00}} & \best{\substd{50.15}{1.09}} & \best{\substd{62.85}{1.47}} \\
    \cmidrule{1-14}
    \multirow{2}{*}{\quad + bagging} 
      & S & \best{\substd{62.00}{0.63}} & $\substd{46.93}{1.17}$ & $\substd{55.24}{0.29}$ & $\substd{21.94}{0.82}$ & \best{\substd{49.60}{0.73}} & $\substd{5.57}{1.72}$ & \best{\substd{54.00}{0.97}} & \best{\substd{67.05}{1.42}} & $\substd{34.95}{0.71}$ & $\substd{19.99}{1.67}$ & \best{\substd{50.18}{1.52}} & $\substd{5.46}{1.31}$ \\
      & C & $\substd{57.26}{0.78}$ & \best{\substd{64.34}{0.79}} & \best{\substd{61.60}{0.52}} & \best{\substd{54.96}{0.73}} & $\substd{48.67}{1.98}$ & \best{\substd{56.28}{2.21}} & $\substd{49.30}{0.56}$ & $\substd{65.49}{1.39}$ & \best{\substd{36.57}{2.79}} & \best{\substd{22.94}{1.90}} & $\substd{49.64}{0.67}$ & \best{\substd{65.17}{1.39}} \\
    \midrule
    \multirow{2}{*}{LLavaMed} 
      & S & \best{\substd{50.26}{0.07}} & $\substd{40.20}{0.27}$ & \best{\substd{50.18}{0.61}} & $\substd{48.56}{0.61}$ & \best{\substd{49.02}{1.08}} & $\substd{40.14}{1.53}$ & $\substd{50.81}{3.99}$ & $\substd{47.09}{4.27}$ & \best{\substd{34.52}{2.71}} & \best{\substd{31.87}{2.87}} & \best{\substd{49.82}{0.47}} & $\substd{42.68}{1.10}$ \\
      & C & $\substd{49.97}{0.57}$ & \best{\substd{45.23}{0.85}} & $\substd{50.13}{0.37}$ & \best{\substd{49.84}{0.36}} & $\substd{47.18}{1.18}$ & \best{\substd{42.73}{1.07}} & \best{\substd{51.54}{2.18}} & \best{\substd{50.13}{1.96}} & $\substd{34.21}{2.67}$ &  $\substd{29.66}{2.53}$ & $\substd{49.07}{1.20}$ & \best{\substd{46.97}{1.96}} \\
    \cmidrule{1-14}
    \multirow{2}{*}{\quad + bagging} 
      & S & $\substd{50.10}{0.82}$ & \best{\substd{50.10}{0.50}} & \best{\substd{50.14}{0.50}} & $\substd{46.75}{0.56}$ & \best{\substd{49.60}{0.74}} & $\substd{35.16}{1.34}$ & $\substd{50.24}{0.81}$ & $\substd{41.66}{1.66}$ & \best{\substd{38.15}{2.86}} & \best{\substd{29.05}{3.55}} & \best{\substd{49.19}{0.71}} & $\substd{36.89}{1.18}$ \\
      & C & \best{\substd{50.15}{0.50}} & $\substd{41.37}{0.73}$ & $\substd{50.14}{0.58}$ & \best{\substd{49.46}{0.60}} & $\substd{45.34}{1.36}$ & \best{\substd{36.79}{1.66}} & \best{\substd{51.70}{1.80}} & \best{\substd{47.59}{2.60}} & $\substd{33.42}{2.16}$ & $\substd{22.56}{2.62}$ & $\substd{48.65}{1.26}$ & \best{\substd{44.29}{1.48}} \\
    \bottomrule
\end{tabular}
}
\end{table*}

\paragraph{Evaluation metrics} We mainly apply the \textit{Balanced Accuracy (BAcc.)}~\citep{brodersen2010balanced} and the \textit{F1 score}~\citep{powers2020evaluation} as our primary evaluation metrics. \textit{BAcc.} computes the average of recall obtained on each class, thereby accounting for the imbalanced distribution by giving equal weight to each class regardless of their frequency. Additionally, we use the \textit{F1} score, which is the harmonic mean of precision and recall, to evaluate the model’s ability to balance false positives and false negatives. The \textit{F1} score is particularly useful in scenarios where high precision or recall alone may not be sufficient to reflect meaningful performance.

To quantify uncertainty, we adopt a \emph{bootstrap-based} statistical testing. For each method and task, we repeatedly resample the test set with replacement and compute BAcc.\ and F1 on each resampled set.  We report the mean and standard deviation of each metric across bootstrap samples, and assess performance differences between \ours{} and single-image baselines using non-parametric bootstrap testing.

\subsection{Off-the-shelf Performance on Medical VLMs}
\label{exp:off-the-shelf}

In this section, we assess whether incorporating reference images influences the \emph{off-the-shelf} diagnostic performance of VLMs. Table~\ref{tab:combined-zs} reports performance for two strategies: (1) the standard single-image input, and (2) the comparative setting (\ours), where the query image is paired with a reference image. For completeness, we additionally report a bagging variant in which each query is evaluated with three independently sampled reference images (see Sec.~\ref{sec:method} for details); these values are listed directly beside each model in the table.

Table~\ref{tab:combined-zs} groups models into general-purpose and medical VLMs. General-purpose VLMs (e.g., QwenVL, Phi) are not trained on medical imaging, which likely contributes to their weaker single-image diagnosis. Medical VLMs vary in specialization: NVILA is pretrained on large-scale radiology/pathology corpora and supports multi-image comparative reasoning~\citep{liu2024nvila}, whereas HuatuoVision and LLaVaMed are primarily trained with single-image objectives across modalities~\citep{chen2024huatuogpt,li2023llava}.


\paragraph{Result analysis}
As shown in Table~\ref{tab:combined-zs}, the effect of \ours{} on off-the-shelf performance is model- and task-dependent. For general-purpose VLMs, introducing a reference image via \ours{} generally improves performance across the majority of tasks. For example, QwenVL and Phi show gains in both BAcc. and F1 for \texttt{Edema} and \texttt{Retinopathy}, despite their weak single-image baselines. For \texttt{Pneumonia} and \texttt{Glaucoma}, at least one of the BAcc. and F1 is increased through the comparative diagnosis. The main exceptions are \texttt{Melanoma} on QwenVL and \texttt{DermaTri} on Phi, plausibly due to limited dermatology priors in general-purpose VLMs: dermoscopic patterns are subtle and distributionally mismatched to natural-image pretraining, leading to consistently weaker performance reported in prior work~\citep{zhou2023pre,zeng2025mm}. As a result, these models cannot reliably interpret lesion morphology, and adding a reference image introduces noise rather than informative contrast. In contrast, abnormalities in other modalities (e.g., lung opacities, retinal lesions) manifest as coarse structural differences that even non-medical VLMs can exploit when given a reference. These improvements indicate that, even without medical priors, general-purpose VLMs \emph{can} benefit from comparative cues when the architecture and pretraining support multi-image inputs.

Medical VLMs exhibit more varied behavior. For NVILA, which is pretrained on large-scale radiology and pathology corpora, we observe a different pattern. On the radiology tasks (\texttt{Pneumonia} and \texttt{Edema}), the single-image setting S already achieves strong BAcc., and \ours{} mainly induces metric-dependent trade-offs: the comparative setting C sometimes increases F1 while slightly reducing BAcc. This suggests that pairing may shift the decision threshold toward more aggressive positive detection, improving precision–recall trade-offs without substantially changing overall discrimination. In contrast, on non-radiology tasks such as \texttt{Glaucoma}, \texttt{Melanoma}, and \texttt{DermaTri}, NVILA’s single-image priors are less aligned with the data distribution, and SiP more clearly improves both BAcc. and F1 by supplying an explicit healthy baseline against which deviations become easier to detect.

HuatuoVision and LLaVaMed exhibit mixed behavior: they can benefit substantially in F1 on some tasks (e.g., \texttt{Glaucoma} and \texttt{Retinopathy}) but also show inconsistent or negative changes in BAcc. on others, indicating that single-image priors and imperfect cross-image integration can interact in a task-dependent way. These models, like NVILA in radiology, are primarily trained on single-image medical datasets. Their representations are therefore biased toward single-view inference and do not naturally support alignment or comparison across images. As a result, the reference image is not effectively integrated into the diagnostic process, and single-image priors dominate prediction behavior. This helps explain why reference images yield mixed and sometimes metric-dependent gains. Taken together, these trends suggest that \ours{} is most beneficial when a model has both (i) sufficiently broad domain priors to interpret the visual features and (ii) an inference mechanism capable of meaningful cross-subject comparison.

\begin{table*}[h!]
\centering
\caption{\textbf{Performance comparison across different data modalities for three VLM backbones on SFT: Qwen-VL-7B, Phi-3-4B, and NVILA-8B.} The table reports BAcc. and F1 scores for six medical datasets among the six columns. Each model is evaluated under various settings (baselines). The highlight of \textbf{bold} and \ul{underline} indicates the \textbf{best} and \ul{second best} performance, respectively. The results show that incorporating~\ours~ consistently improves performance. Thus, \ours~provides a simple and effective strategy to improve VLM performance.}
\resizebox{0.99\linewidth}{!}{%
\begin{tabular}{l|
  *{2}{c} |
  *{2}{c} |
  *{2}{c} |
  *{2}{c} |
  *{2}{c} |
  *{2}{c}}
\toprule
\multirow{2}{*}{\textbf{Settings}} &
\multicolumn{2}{c}{\textbf{\textbf{Pneumonia}}} &
\multicolumn{2}{c}{\textbf{\textbf{Edema}}} &
\multicolumn{2}{c}{\textbf{\textbf{Glaucoma}}} &
\multicolumn{2}{c}{\textbf{\textbf{Melanoma}}} &
\multicolumn{2}{c}{\textbf{\textbf{DermaTri}}} &
\multicolumn{2}{c}{\textbf{\textbf{Retinopathy}}} \\
\cmidrule(lr){2-3}\cmidrule(lr){4-5}\cmidrule(lr){6-7}\cmidrule(lr){8-9}\cmidrule(lr){10-11}\cmidrule(lr){12-13}
 & BAcc. & F1 
 & BAcc. & F1 
 & BAcc. & F1 
 & BAcc. & F1 
 & BAcc. & F1 
 & BAcc. & F1 \\
\midrule
\multicolumn{13}{c}{\textbf{Qwen-VL-7B}}\\
\textsc{Single (All)} & $\substd{50.00}{0.00}$ & $\substd{33.34}{0.36}$  & $\substd{50.00}{0.00}$ & $\substd{33.32}{0.25}$  & $\substd{50.00}{0.00}$ & $\substd{32.86}{0.92}$  & $\substd{57.09}{0.90}$ & $\substd{48.50}{1.57}$  & $\substd{62.08}{1.77}$ & $\substd{55.45}{2.48}$  & $\substd{50.00}{0.00}$ & $\substd{33.32}{0.66}$ \\

\textsc{Single (Rand)}  & $\substd{64.09}{0.72}$ & $\substd{63.44}{0.76}$  & $\substd{74.01}{0.49}$ & $\substd{73.97}{0.49}$  & $\substd{64.42}{1.93}$ & $\substd{64.17}{1.98}$  & $\substd{73.55}{1.38}$ & $\substd{73.50}{1.39}$  & $\substd{65.24}{2.36}$ & $\substd{64.70}{2.74}$  & $\substd{68.46}{1.19}$ & $\substd{66.29}{1.48}$ \\

\textsc{Single (Cluster)}  & $\substd{56.99}{0.48}$ & $\substd{49.27}{0.79}$  & $\substd{73.04}{0.47}$ & $\substd{72.51}{0.51}$  & $\substd{54.03}{1.41}$ & $\substd{48.03}{1.98}$  & $\substd{64.87}{1.27}$ & $\substd{62.55}{1.54}$  & $\substd{63.02}{2.32}$ & $\substd{60.45}{2.78}$  & $\substd{69.76}{1.22}$ & $\substd{68.44}{1.40}$ \\

\textsc{Single (Spatial)}  & $\substd{64.12}{0.68}$ & $\substd{62.10}{0.81}$  & $\substd{69.52}{0.48}$ & $\substd{68.62}{0.52}$  & $\substd{64.43}{1.91}$ & $\substd{63.80}{1.99}$  & $\substd{68.80}{1.39}$ & $\substd{67.84}{1.52}$  & $\substd{53.29}{1.82}$ & $\substd{43.92}{1.86}$  & $\substd{73.93}{1.23}$ & $\substd{73.32}{1.31}$ \\

\textsc{Single (Coverage)}  & $\substd{68.25}{0.73}$ & $\substd{68.08}{0.75}$  & $\substd{74.45}{0.47}$ & $\substd{73.94}{0.51}$  & $\substd{62.66}{2.01}$ & $\substd{62.47}{2.02}$  & $\substd{64.36}{1.30}$ & $\substd{61.95}{1.58}$  & $\substd{66.91}{2.57}$ & $\substd{66.69}{2.73}$  & $\substd{74.43}{1.30}$ & $\substd{74.40}{1.31}$ \\

\textsc{Single (Bagging)}  & $\substd{64.75}{0.68}$ & $\substd{63.44}{0.77}$  & $\substd{73.75}{0.49}$ & $\substd{73.69}{0.49}$  & $\substd{64.15}{1.98}$ & $\substd{64.01}{2.01}$  & $\substd{71.70}{1.28}$ & $\substd{70.83}{1.43}$  & $\substd{62.46}{2.26}$ & $\substd{59.68}{2.78}$  & $\substd{72.39}{1.22}$ & $\substd{71.48}{1.37}$ \\

\hline

\ours  & \secondbest{\substd{68.94}{0.74}} & \secondbest{\substd{68.83}{0.74}}  & $\substd{77.79}{0.46}$ & $\substd{77.73}{0.47}$  & \secondbest{\substd{65.87}{1.91}} & \secondbest{\substd{65.45}{2.00}}  & \best{\substd{79.44}{1.24}} & \best{\substd{79.32}{1.27}}  & \best{\substd{79.39}{2.20}} & \best{\substd{79.04}{2.28}}  & $\substd{79.38}{1.21}$ & $\substd{79.36}{1.21}$ \\

\ours~\textsc{(w/ bagging)}  & $\substd{68.68}{0.69}$ & $\substd{68.56}{0.70}$  & \secondbest{\substd{77.88}{0.45}} & \secondbest{\substd{77.81}{0.46}}  & $\substd{65.41}{1.91}$ & $\substd{65.02}{1.99}$  & \secondbest{\substd{79.30}{1.29}} & \secondbest{\substd{79.16}{1.33}}  & \secondbest{\substd{78.96}{2.17}} & \secondbest{\substd{78.74}{2.22}}  & \secondbest{\substd{79.41}{1.26}} & \secondbest{\substd{79.41}{1.26}} \\

\ours~\textsc{(w/ embedding)} & \best{\substd{69.09}{0.74}} & \best{\substd{68.96}{0.75}}  & \best{\substd{78.18}{0.46}} & \best{\substd{78.12}{0.46}}  & \best{\substd{66.44}{1.92}} & \best{\substd{66.34}{1.92}}  & $\substd{78.84}{1.26}$ & $\substd{78.63}{1.31}$  & $\substd{77.45}{2.28}$ & $\substd{77.09}{2.34}$ & \best{\substd{79.56}{1.21}} & \best{\substd{79.54}{1.21}} \\

\midrule

\multicolumn{13}{c}{\textbf{\textbf{Phi-3-4B}}}\\
\textsc{Single (All)} & $\substd{50.00}{0.00}$ & $\substd{33.32}{0.36}$  & $\substd{50.00}{0.00}$ & $\substd{33.33}{0.25}$  & $\substd{52.66}{0.75}$ & $\substd{39.27}{1.64}$  & $\substd{51.82}{0.65}$ & $\substd{38.96}{1.28}$  & $\substd{62.38}{1.59}$ & $\substd{54.64}{2.30}$  & $\substd{50.00}{0.00}$ & $\substd{33.32}{0.66}$ \\

\textsc{Single (Rand)} & $\substd{59.26}{0.73}$ & $\substd{57.59}{0.81}$  & $\substd{58.27}{0.39}$ & $\substd{51.97}{0.58}$  & $\substd{51.56}{0.61}$ & $\substd{37.63}{1.46}$  & $\substd{58.61}{1.00}$ & $\substd{51.82}{1.58}$  & $\substd{68.25}{2.43}$ & $\substd{67.17}{2.67}$  & $\substd{71.21}{1.41}$ & $\substd{71.12}{1.43}$ \\

\textsc{Single (Cluster)} & $\substd{52.30}{0.30}$ & $\substd{39.28}{0.62}$  & \secondbest{\substd{64.75}{0.47}} & \secondbest{\substd{62.94}{0.55}}  & $\substd{64.52}{1.79}$ & $\substd{64.30}{1.94}$  & $\substd{51.71}{0.43}$ & $\substd{37.22}{1.12}$  & $\substd{72.98}{2.38}$ & $\substd{72.31}{2.57}$  & $\substd{71.21}{1.40}$ & $\substd{71.19}{1.40}$ \\

\textsc{Single (Spatial)} & $\substd{56.01}{0.50}$ & $\substd{48.49}{0.79}$  & $\substd{55.17}{0.32}$ & $\substd{46.16}{0.54}$  & $\substd{64.46}{1.83}$ & $\substd{64.57}{1.94}$  & $\substd{51.02}{0.33}$ & $\substd{35.59}{1.01}$  & $\substd{72.08}{2.36}$ & $\substd{71.22}{2.59}$  & $\substd{70.30}{1.41}$ & $\substd{70.04}{1.43}$ \\

\textsc{Single (Coverage)} & $\substd{61.58}{0.67}$ & $\substd{58.78}{0.83}$  & $\substd{61.60}{0.45}$ & $\substd{57.97}{0.57}$  & $\substd{64.35}{1.85}$ & $\substd{63.04}{2.01}$  & $\substd{51.81}{0.44}$ & $\substd{37.42}{1.14}$  & $\substd{71.03}{2.39}$ & $\substd{70.15}{2.62}$  & $\substd{67.33}{1.30}$ & $\substd{65.88}{1.49}$ \\

\textsc{Single (Bagging)}  & $\substd{57.90}{0.57}$ & $\substd{52.65}{0.80}$ & $\substd{64.19}{0.47}$ & $\substd{61.61}{0.59}$ & $\substd{56.72}{1.06}$ & $\substd{47.46}{1.90}$  & $\substd{55.30}{0.70}$ & $\substd{44.63}{1.42}$ & $\substd{70.22}{2.38}$ & $\substd{69.21}{2.62}$ & $\substd{71.76}{1.38}$ & $\substd{71.76}{1.38}$ \\

\hline

\ours & \secondbest{\substd{63.08}{0.72}} & \secondbest{\substd{62.05}{0.77}}  & $\substd{63.82}{0.44}$ & $\substd{60.46}{0.58}$  & \secondbest{\substd{65.07}{1.97}} & \secondbest{\substd{64.94}{2.00}}  & \secondbest{\substd{65.98}{1.30}} & \secondbest{\substd{64.01}{1.55}}  & $\substd{68.94}{2.00}$ & $\substd{65.14}{2.66}$  & \secondbest{\substd{73.88}{1.30}} & \secondbest{\substd{73.53}{1.34}} \\

\ours~\textsc{(w/ bagging)} & $\substd{62.62}{0.70}$ & $\substd{61.43}{0.77}$  & $\substd{63.71}{0.09}$ & $\substd{60.31}{0.13}$  & \best{\substd{65.17}{1.97}} & \best{\substd{65.05}{2.00}}  & \best{\substd{66.80}{1.27}} & \best{\substd{64.98}{1.51}}  & \secondbest{\substd{75.73}{2.13}} & \secondbest{\substd{75.09}{2.35}}  & $\substd{73.83}{1.26}$ & $\substd{73.51}{1.33}$ \\

\ours~\textsc{(w/ embedding)} & \best{\substd{63.71}{0.72}} & \best{\substd{62.68}{0.78}}  & \best{\substd{66.05}{0.44}} & \best{\substd{63.29}{0.57}}  & $\substd{63.14}{1.96}$ & $\substd{63.05}{1.97}$  & $\substd{65.20}{1.29}$ & $\substd{62.77}{1.58}$  & \best{\substd{76.79}{2.23}} & \best{\substd{76.20}{2.44}} & \best{\substd{74.03}{1.29}} & \best{\substd{73.72}{1.33}} \\

\midrule

\multicolumn{13}{c}{\textbf{NVILA-8B}}\\
\textsc{Single (All)} & $\substd{50.28}{0.09}$ & $\substd{34.00}{0.41}$  & $\substd{51.68}{0.15}$ & $\substd{37.14}{0.39}$  & $\substd{58.98}{1.31}$ & $\substd{52.09}{2.06}$  & $\substd{43.09}{0.87}$ & $\substd{31.48}{0.96}$  & $\substd{62.42}{1.96}$ & $\substd{58.46}{2.65}$  & $\substd{51.42}{0.38}$ & $\substd{36.55}{1.02}$ \\

\textsc{Single (Rand)} & $\substd{79.08}{0.64}$ & $\substd{79.05}{0.65}$  & $\substd{84.74}{0.40}$ & $\substd{84.65}{0.41}$  & $\substd{60.19}{1.89}$ & $\substd{60.06}{1.91}$ & $\substd{69.14}{1.42}$ & $\substd{68.70}{1.48}$ & $\substd{74.38}{2.22}$ & $\substd{73.52}{2.50}$ & $\substd{76.66}{1.24}$ & $\substd{76.29}{1.33}$ \\

\textsc{Single (Cluster)} & $\substd{76.64}{0.64}$ & $\substd{76.27}{0.68}$  & $\substd{85.13}{0.39}$ & $\substd{85.05}{0.40}$  & $\substd{63.52}{1.91}$ & $\substd{63.47}{1.91}$  & $\substd{74.30}{1.36}$ & $\substd{74.25}{1.37}$  & $\substd{78.23}{2.24}$ & $\substd{78.34}{2.21}$  & $\substd{73.52}{1.18}$ & $\substd{72.41}{1.38}$ \\

\textsc{Single (Spatial)} & $\substd{79.23}{0.64}$ & $\substd{79.22}{0.64}$  & $\substd{84.97}{0.39}$ & $\substd{84.85}{0.41}$  & $\substd{62.91}{1.91}$ & $\substd{62.81}{1.91}$  & $\substd{74.09}{1.31}$ & $\substd{73.88}{1.36}$  & $\substd{77.34}{2.26}$ & $\substd{77.51}{2.23}$  & $\substd{79.92}{1.20}$ & $\substd{79.78}{1.23}$ \\

\textsc{Single (Coverage)} & $\substd{71.71}{0.62}$ & $\substd{70.21}{0.74}$  & $\substd{78.72}{0.42}$ & $\substd{77.91}{0.50}$  & $\substd{49.90}{0.68}$ & $\substd{36.03}{1.32}$  & $\substd{73.56}{1.37}$ & $\substd{73.48}{1.38}$  & $\substd{77.33}{2.28}$ & $\substd{77.47}{2.25}$  & $\substd{77.31}{1.21}$ & $\substd{76.88}{1.31}$ \\

\textsc{Single (Bagging)}  & $\substd{78.75}{0.66}$ & $\substd{78.66}{0.67}$  & $\substd{84.65}{0.40}$ & $\substd{84.55}{0.41}$  & $\substd{62.01}{1.97}$ & $\substd{61.94}{1.97}$  & $\substd{74.20}{1.38}$ & $\substd{74.20}{1.38}$  & $\substd{77.99}{2.29}$ & $\substd{78.02}{2.28}$  & $\substd{78.69}{1.17}$ & $\substd{78.47}{1.22}$ \\

\hline

\ours & \secondbest{\substd{79.80}{0.66}} & \secondbest{\substd{79.79}{0.66}}  & $\substd{85.27}{0.40}$ & $\substd{85.25}{0.40}$  & \best{\substd{69.63}{1.82}} & \best{\substd{68.96}{1.93}}  & \secondbest{\substd{76.36}{1.30}} & \secondbest{\substd{76.33}{1.31}}  & \secondbest{\substd{81.17}{2.13}} & \secondbest{\substd{81.03}{2.16}}  & \secondbest{\substd{81.57}{1.14}} & $\substd{81.54}{1.15}$ \\

\ours~\textsc{(w/ bagging)} & \best{\substd{80.05}{0.65}} & \best{\substd{80.04}{0.65}}  & \secondbest{\substd{85.46}{0.40}} & \secondbest{\substd{85.45}{0.40}}  & \secondbest{\substd{69.13}{1.87}} & \secondbest{\substd{68.38}{1.97}}  & \best{\substd{76.80}{1.35}} & \best{\substd{76.78}{1.35}}  & $\substd{80.58}{2.16}$ & $\substd{80.50}{2.19}$ & $\substd{81.56}{1.16}$ & \secondbest{\substd{81.55}{1.17}} \\

\ours~\textsc{(w/ embedding)} & $\substd{79.56}{0.64}$ & $\substd{79.55}{0.64}$  & \best{\substd{85.92}{0.39}} & \best{\substd{85.90}{0.39}}  & $\substd{68.05}{1.80}$ & $\substd{67.00}{1.96}$  & $\substd{75.91}{1.33}$ & $\substd{75.84}{1.34}$  & \best{\substd{81.21}{2.15}} & \best{\substd{81.15}{2.15}} & \best{\substd{81.73}{1.15}} & \best{\substd{81.71}{1.16}} \\
\bottomrule
\end{tabular}%
}
\label{tab:baseline}
\end{table*}

\subsection{Comparison to Baselines in SFT Settings}
\label{exp:sft}
The off-the-shelf results reveal two gaps: missing medical knowledge and missing comparison capability. Here we ask: \emph{Can lightweight SFT with paired (query, reference) inputs (\ours) outperform single-image SFT baselines?}

\paragraph{Settings}
To isolate the effect of \emph{comparison structure} from data quantity, we first hold out the test set to prevent leakage during both training and reference selection.
From the remaining split, we fix a \emph{shared disease-query set} across all methods: for binary tasks, 500 positive (diseased) query images; for the multi-class task (\texttt{DermaTri}), 411 query images per disease class (822 total).
\emph{Negatives are the experimental variable:} baselines differ only in (i) how negative images are selected from the negative pool, and (ii) whether negatives are used as standard single-image training samples or as paired healthy-control references for cross-subject comparison.
Unless otherwise stated, each size-matched baseline uses 500 negatives for binary tasks and 211 negatives for \texttt{DermaTri}. For VLMs, we compare two general-purpose models (QwenVL and Phi) and one model with medical priors (NVILA) under matched prompting and adaptation settings to isolate the effect of reference-conditioned comparative diagnosis. Many specialized medical VLMs (e.g., LLaVA-Med) are primarily optimized for single-image, closed-set prediction; adapting them to our paired-input protocol would require substantial re-engineering (e.g., paired training objectives and modality-specific preprocessing pipelines), reducing comparability and hindering reproducibility.

\paragraph{Baselines}
With the shared disease-query set fixed, we compare \ours{} against a size-matched family of single-image SFT baselines:
\begin{itemize}
    \item \textsc{Single (Rand)}: randomly sample 500 negatives per task.
    \item \textsc{Single (Cluster)}: select 500 negatives via embedding-space clustering to encourage diversity~\citep{pratap2023novel}.
    \item \textsc{Single (Spatial)}: select 500 negatives using spatial/instance-aware criteria to reduce redundancy~\citet{yan2022spatial}.
    \item \textsc{Single (Coverage)}: select 500 negatives to maximize coverage of the negative distribution~\citet{wang2025resampling}.
    \item \textsc{Single (Bagging)}: ensemble multiple \textsc{Single (Rand)} models trained on independently resampled 500-negative subsets; aggregate by majority vote.
    \item \textsc{Single (All)}: use the entire available negative pool (more than 500 negatives) as an upper bound for ``more negatives'' without pairing.
\end{itemize}

\paragraph{Implementation}
All methods use the same lightweight SFT configuration with LoRA adapters (rank 16), trained with the Hugging Face Trainer\footnote{\url{https://huggingface.co/docs/transformers/en/main_classes/trainer}}
and Accelerate\footnote{\url{https://huggingface.co/docs/accelerate/package_reference/accelerator}}
on four A100 GPUs (per-device batch size 1, learning rate \(1\times10^{-4}\)).
For fairness, we fix the total number of optimizer updates across all settings by adjusting gradient accumulation to match the effective dataset size.
When \ours{} uses multiple references per query, we average gradients across the paired tuples and take a single optimizer step, preserving the same update budget. Our code will be open-sourced upon acceptance.

\paragraph{Result analysis}
Table~\ref{tab:baseline} reports SFT performance across three VLM backbones and six diagnostic tasks. Across nearly all modalities and architectures, some variant of \ours{} (\ours{}, \textsc{\ours{} (w/ bagging)}, or \textsc{\ours{} (w/ embedding)}) is consistently competitive and typically achieves best or second-best performance. For Qwen-VL, this lightweight comparative tuning yields improvements on every task. Phi-3-4B, despite its limited medical priors, benefits even more from \ours{}, especially in dermatology and ophthalmology settings where single-image cues are weaker. 
NVILA-8B, whose radiology-focused pretraining already delivers strong single-image performance on \texttt{Pneumonia} and \texttt{Edema}, still gains markedly when \ours{} is applied, indicating that comparative diagnosis provides transferable structure beyond its original training domain. Even when compared against the curated single-image strategies (\textsc{Single (Cluster)}, \textsc{Single (Spatial)}, \textsc{Single (Coverage)}) and the bagged ensemble, \ours{} almost always attains the best or second-best BAcc./F1 scores. 
Overall, gains are particularly pronounced in ophthalmology (\texttt{Glaucoma}, \texttt{Retinopathy}) and dermatology (\texttt{Melanoma}, \texttt{DermaTri}), where reliable diagnosis hinges on subtle morphological differences: precisely the type of signal that comparative supervision exposes during SFT (see further analysis in Sec.~\ref{exp:heatmap} and Sec.~\ref{sec:theory}).

\begin{table}[t]
\centering
\caption{\textbf{Performance and alignment across reference-matching criteria.}
Reported metrics include balanced accuracy (BAcc.), F1 score, Cohen's $\kappa$,
and agreement (percentage of matching predictions) for pneumonia and edema classification.}
\label{tab:ref-matching}
\setlength{\tabcolsep}{3pt}  
\renewcommand{\arraystretch}{1.2}
\footnotesize
\begin{tabular}{@{}lcccccccc@{}}
\toprule
& \multicolumn{4}{c}{\textbf{Pneumonia}} & \multicolumn{4}{c}{\textbf{Edema}} \\
\cmidrule(lr){2-5}\cmidrule(lr){6-9}
\textbf{Strategy\textsuperscript{\textit{a}}}
  & \textbf{BAcc.} & \textbf{F1} & \textbf{$\kappa$} & \textbf{Agree}
  & \textbf{BAcc.} & \textbf{F1} & \textbf{$\kappa$} & \textbf{Agree} \\
\midrule
\rowcolor{gray!8}
SVP & 68.92 & 66.99 & 92.97 & 96.53 & 77.80 & 78.94 & 97.45 & 98.74 \\
SV  & 68.55 & 66.61 & 93.52 & 96.80 & 77.80 & 78.88 & 97.65 & 98.84 \\
\rowcolor{gray!8}
SP  & 68.92 & 66.69 & 92.97 & 96.53 & 77.80 & 78.94 & 97.45 & 98.74 \\
VP  & 68.70 & 66.77 & 92.51 & 96.30 & 77.82 & 78.97 & 97.70 & 98.86 \\
\rowcolor{gray!8}
S   & 68.78 & 67.11 & 92.27 & 96.17 & 77.72 & 78.82 & 97.25 & 98.64 \\
V   & 68.63 & 66.70 & 92.16 & 96.13 & 78.05 & 79.13 & 97.40 & 98.71 \\
\rowcolor{gray!8}
P   & 68.70 & 66.77 & 92.51 & 96.30 & 77.82 & 78.97 & 97.70 & 98.62 \\
\midrule
CC  & 68.43 & 67.15 & 91.58 & 95.83 & 77.90 & 79.04 & 97.24 & 98.64 \\
\rowcolor{gray!8}
EB  & 69.07 & 66.97 & 92.15 & 96.13 & 78.19 & 79.29 & 97.32 & 98.68 \\
RS  & 68.70 & 67.05 & \multicolumn{1}{c}{---} & \multicolumn{1}{c}{---}
    & 78.04 & 79.10 & \multicolumn{1}{c}{---} & \multicolumn{1}{c}{---} \\
\bottomrule
\end{tabular}

\vspace{2pt}
\raggedright
\scriptsize
\textsuperscript{\textit{a}}Strategy abbreviations:
SVP (sex + views + projection),
SV (sex + frontal views),
SP (sex + projection),
VP (views + projection),
S (sex),
V (views),
P (projection),
CC (cross-center),
EB (embedding-based top-1 retrieval),
RS (random sampling).
\end{table}

\subsection{Selection Strategies}
\label{exp:selection}
A practical concern for clinical deployment is whether \ours{} requires carefully matched references. If performance depends on tight demographic or acquisition matching, the method's applicability would be limited. We test this by systematically relaxing matching constraints.

\paragraph{Settings}
We conduct this analysis on CheXpert~\citep{irvin2019chexpert} for two reasons. \emph{First}, CheXpert provides rich and standardized metadata, including patient sex, view position, projection, and other acquisition descriptors. This demographic information enables us to construct reference sets under progressively weaker matching constraints in a controlled and reproducible manner. Other imaging datasets in our benchmark do not contain sufficiently detailed
metadata to support the full set of ablations performed here. \emph{Second}, CheXpert has a natural companion dataset in MIMIC-CXR~\citep{johnson2019mimic}, a large radiology collection acquired at a different clinical center but annotated with a closely aligned metadata schema. This allows us to evaluate cross-center robustness by drawing references from a different institution while maintaining comparable metadata fields. 

Using the \texttt{Pneumonia} and \texttt{Edema} tasks in CheXpert, we construct matched reference pools using 
nine selection criteria. The first seven impose progressively weaker metadata-based 
constraints:
\emph{S} (sex),
\emph{V} (views), \emph{P} (projection),
\emph{SV} (sex + views),
\emph{SP} (sex + projection),
\emph{VP} (views + projection), \emph{SVP} (sex + views + projection).
We additionally include two variants designed to probe robustness beyond demographic and clinical-based matching.
\emph{CC} (\textbf{cross-center sampling}) selects reference images from a different dataset (MIMIC-CXR~\citep{johnson2019mimic}) than the evaluation set (CheXpert), allowing us to test whether the method relies on within-domain reference distributions. There is also an 
\emph{EB} (\textbf{embedding-based similarity}) strategy, which constructs reference sets by retrieving the most similar image in a high-dimensional visual embedding space (we use BioMedCLIP~\citep{zhang2023biomedclip} to extract the embeddings), thereby removing explicit metadata constraints and simulating a content-based retrieval scenario.
Finally, \emph{RS} (random sampling) serves as a baseline without any matching constraints.
Together, these conditions evaluate whether \ours{} requires tightly matched references or remains stable under substantially weaker, cross-domain, or purely similarity-based reference selection.

\begin{figure*}[h]
    \centering
    \includegraphics[width=\linewidth]{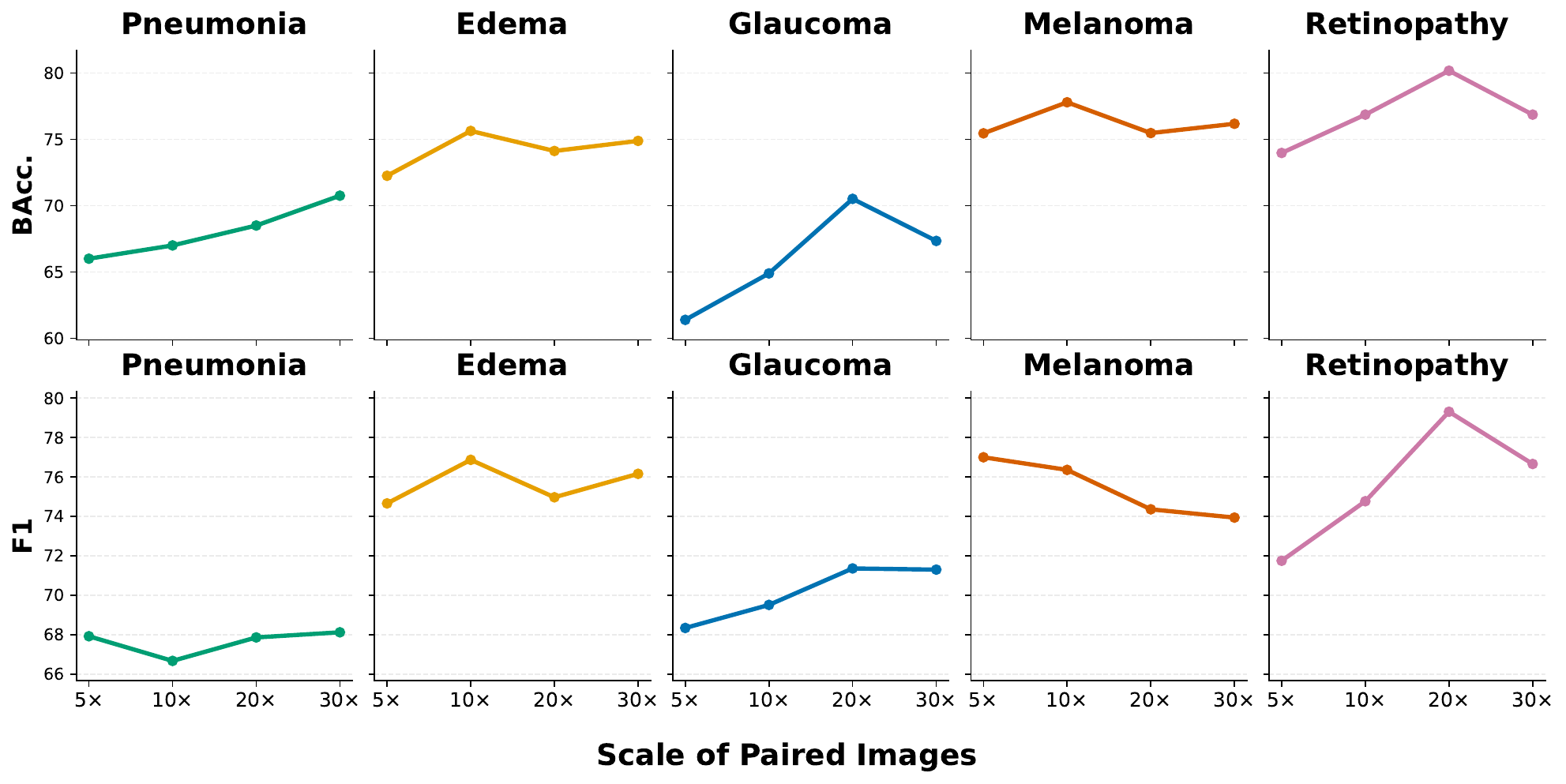}
    \caption{\textbf{Impact of the number of reference images.} Each column corresponds to one data modality (\texttt{Pneumonia}, \texttt{Edema}, \texttt{Glaucoma}, \texttt{Melanoma}, \texttt{Retinopathy}). The \(x\)-axis shows the ratio \(K\) of reference images to each query image (\(5\times,10\times,20\times,30\times\)). The top row reports balanced accuracy (BAcc.) and the bottom row reports F1.}
    \label{fig:num-ref}
\end{figure*}

\paragraph{Evaluation metrics}
Beyond BAcc. and F1 score, we report two agreement measures that 
quantify the consistency between a given reference-selection strategy and the random 
baseline (RS). Cohen’s~$\kappa$ captures chance-corrected agreement~\citep{cohen1960coefficient}, where values near~100 
indicate almost perfect consistency. Pairwise accuracy agreement reports the percentage 
of samples on which two models produce identical predictions, providing an interpretable 
measure of behavioral similarity. 


\paragraph{Results}
Table~\ref{tab:ref-matching} reports results on the \texttt{Pneumonia} and 
\texttt{Edema} performance across all nine matching strategies. Performance remains stable as metadata constraints are relaxed: moving from SVP to S, V, or P yields fluctuations within about one percentage point in BAcc. and F1, suggesting limited sensitivity to tightly aligned patient attributes or acquisition conditions within the tested ranges. Performance also remains strong under \emph{cross-center} 
sampling (CC) and \emph{embedding-based} retrieval (EB), demonstrating robustness even 
when references come from a different institution or are selected purely based on visual 
similarity.

Agreement metrics mirror this behavior. Cohen’s~$\kappa$ remains high 
(around $0.90$--$0.97$), pairwise agreement consistently exceeds 95-98\%. Collectively, these findings show that once trained with \ours{}, the 
model extracts pathology-relevant contrasts in a way that is largely insensitive to how 
the reference image is chosen. This robustness is crucial for real-world deployment, 
where metadata completeness and domain alignment cannot always be guaranteed.

Having found \ours{} to be relatively insensitive to \emph{which} reference images are selected under these strategies, we next study how performance depends on \emph{how many} references are available during training.

\subsection{Influence on the Scale of Reference Images}
\label{exp:scale}
\ours{} constructs comparative SFT tuples by pairing a query image $X$ with a matched reference image $X'$. In practice, any single reference can be imperfect and biased, which increases variance in the training signal. To stabilize optimization, SiP can sample multiple references for the same query and update the model using the \emph{averaged gradient} over these tuples as detailed in Sec.~\ref{sec:method}. We therefore study how the number of references per query affects performance and robustness.

\paragraph{Settings}

Intuitively, for each query $X$ we sample $K$ matched reference images $X'$ and construct comparative SFT inputs $(X, X', Q, A)$. We vary $K$ (shown as $5\times$, $10\times$, $20\times$, and $30\times$ in Fig.~\ref{fig:num-ref}). During training, when multiple references are sampled for the same $X$, we update the model using the averaged gradient across the $K$ tuples, reducing the influence of any single reference choice.

\paragraph{Results}
Fig.~\ref{fig:num-ref} shows that in some cases (e.g., \texttt{Pneumonia}, \texttt{Edema}, and \texttt{Retinopathy}, increasing $K$ slightly improves performance from $5\times$ to $20$--$30\times$, suggesting that multi-reference training strengthens the comparative supervision signal and reduces sensitivity to noisy pairings. The gains exhibit diminishing returns at larger $K$, and some tasks (e.g., \texttt{Glaucoma} and \texttt{Melanoma}) show saturation or slight drops, indicating that very large reference sets may introduce weaker comparisons that add limited new information. Overall, these results support multi-reference training with averaged gradients as a simple and effective way to stabilize \ours{} and improve performance.


\subsection{Qualitative Analysis}
\label{exp:heatmap}
\begin{figure*}
    \centering
    \includegraphics[width=0.9\linewidth]{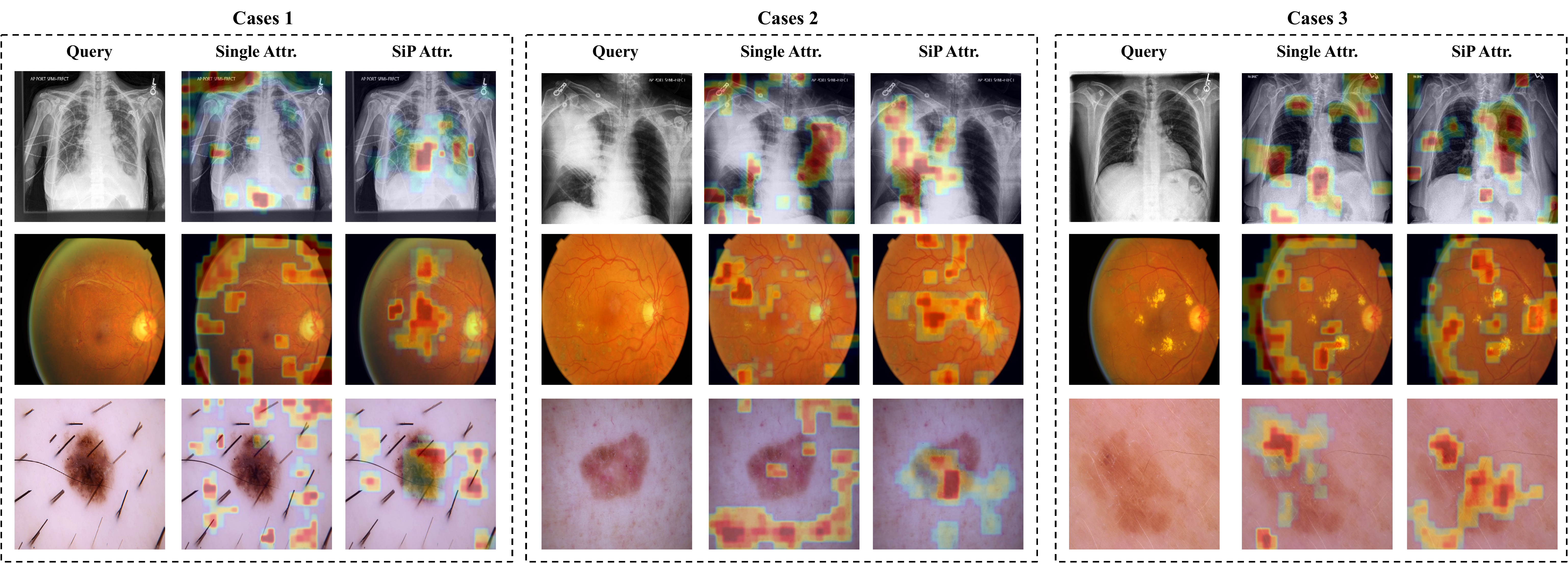}
    \caption{{\textbf{Attribution maps for single-image inference vs.\ SiP.}
    We visualize pixel-level attributions on the \textit{query image} for three representative example sets (Case~1--3) across three modalities (top-to-bottom: \textit{chest X-ray}, \textit{fundus photography}, \textit{dermoscopy}). For each query image (\textit{Query}), we overlay-sensitivity attributions for a single-image SFT (\textit{Single Attr.}) and a \ours{} comparison SFT (\textit{SiP Attr.}). For \ours, the model receives the query together with a matched healthy-control reference image and an explicit comparison instruction.
\textit{Color encodes attribution strength:} warmer colors (yellow/red) indicate higher attribution magnitude (greater sensitivity to occlusion), while cooler colors (blue/green) indicate weaker attribution.}
}
    \label{fig:heatmap}
\end{figure*}

\paragraph{Settings}
To better understand how \ours{} alters the visual grounding of a VLM, we conduct a qualitative attribution study comparing \textit{(1) single-image inference} and (2) \ours, where the query is paired with a \textit{healthy-control reference} and prompted with an explicit comparison instruction.
Concretely, we use two QwenVL variants trained under our SFT protocol: a \textit{single-image SFT model} and a \textit{\ours{}-based SFT model}.
Our objective is to test whether comparative conditioning changes \emph{where} the model looks when making a closed-ended diagnosis.

We generate pixel-level attribution heatmaps using \emph{Occlusion Sensitivity} in Captum~\citep{kokhlikyan2020captum}, which measures the change in the model’s decision when local image regions are masked. Images are resized to $336\times336$, and occlusion is applied using a $32\times32$ sliding window with stride $16$. For \ours, occlusion is performed with both the query and reference images present in the input, and we report attributions for the \emph{query image} in Fig.~\ref{fig:heatmap}. Gaussian smoothing and robust re-normalization are applied as post-processing to reduce block artifacts and improve spatial coherence.

\paragraph{Results}
As shown in Fig.~\ref{fig:heatmap}, the \textit{single-image} setting often yields patchy, spuriously correlated, and weakly structured attribution patterns: highlighted regions appear as scattered blocks and can cover large portions of the image, suggesting reliance on coarse, global evidence rather than specific pathology cues. This is particularly visible in modalities with strong global appearance variation (e.g., fundus illumination and dermoscopy), where attribution can spread broadly instead of aligning to a small set of diagnostically meaningful regions.

In contrast, \ours{} produces more spatially coherent and anatomically plausible saliency across the same query images. The \ours{} overlays tend to (i) reduce spurious activations, (ii) concentrate attribution into fewer, higher-intensity regions, and (iii) better respect the modality-specific structure. For example, in \textit{chest X-rays}, \ours{} more consistently emphasizes regions within the lung fields instead of distributing attribution across the background or side markers. In \textit{fundus images}, \ours{} yields more localized hotspots on retinal regions rather than diffuse activation over the entire fundus. In \textit{dermoscopy}, attribution under \ours{} shifts toward the lesion interior and borders while attenuating attention on surrounding skin and acquisition artifacts. 

Overall, these visualizations suggest that \ours{} reshapes the model's decision process from ``recognize a class from one image'' toward ``identify \emph{where the query deviates from a healthy control}.''
This qualitatively supports the empirical gains reported earlier. In the next section, we deep dive into the underlying mechanism of how \ours{} achieve this effect. 
\section{Conclusion}
This work underscores the value of reference image-guided comparative reasoning in enhancing medical image diagnosis using VLMs. Drawing direct inspiration from clinical practices, we show that incorporating healthy control reference images enables general-purpose VLMs to outperform traditional single-image baselines across a variety of imaging modalities. Our proposed ``See-in-Pairs'' (\ours) framework offers a lightweight yet effective SFT strategy that not only boosts diagnostic accuracy but also improves feature compactness and sample efficiency. Theoretical analysis further supports our findings, demonstrating improved statistical convergence when using paired representations. Altogether, our study provides compelling empirical and theoretical evidence for embracing comparative inference in future medical AI systems, advocating for a paradigm shift toward more clinically-aligned and interpretable VLM architectures.

\section*{Acknowledgments}
This work is supported in part by the Natural Sciences and Engineering Research Council of Canada (NSERC), Public Safety Canada (NS-5001-22170), National Key R\&D Program of China Grant
2024YFA1015800 and NVIDIA Hardware Award.





\bibliographystyle{alpha}
\bibliography{refs}

\end{document}